# U-Net-Like Spiking Neural Networks for Single Image Dehazing

Huibin Li
*College of Computer Science and Cyber Security,*
*(Chengdu University of Technology)*
*School of Data Science and Artificial Intelligence*
*(Wenzhou University of Technology)*
Chengdu, Wenzhou, China
lihuibin@stu.cdut.edu.cn

Haoran Liu
*College of Nuclear Technology and Automation Engineering*
*(Chengdu University of Technology)*
*School of Data Science and Artificial Intelligence*
*(Wenzhou University of Technology)*
Chengdu, Wenzhou, China
liuhaoran@cdut.edu.cn

Mingzhe Liu*
*School of Data Science and Artificial Intelligence*
*(Wenzhou University of Technology)*
Wenzhou, China
liumz@cdut.edu.cn

Yulong Xiao
*Department of Energy*
*(Politecnico di Milano)*
Milano, Italy
yulong.xiao@mail.polimi.it

Peng Li
*College of Nuclear Technology and Automation Engineering*
*(Chengdu University of Technology)*
Chengdu, China
lipeng@stu.cdut.edu.cn

Guibin Zan*
*Sigray, Inc.*
Concord, USA
gbzan@sigray.com

*Abstract*—Image dehazing is a critical challenge in computer vision, essential for enhancing image clarity in hazy conditions. Traditional methods often rely on atmospheric scattering models, while recent deep learning techniques, specifically Convolutional Neural Networks (CNNs) and Transformers, have improved performance by effectively analyzing image features. However, CNNs struggle with long-range dependencies, and Transformers demand significant computational resources. To address these limitations, we propose DehazeSNN, an innovative architecture that integrates a U-Net-like design with Spiking Neural Networks (SNNs). DehazeSNN captures multi-scale image features while efficiently managing local and long-range dependencies. The introduction of the Orthogonal Leaky-Integrate-and-Fire Block (OLIFBlock) enhances cross-channel communication, resulting in superior dehazing performance with reduced computational burden. Our extensive experiments show that DehazeSNN is highly competitive to state-of-the-art methods on benchmark datasets, delivering high-quality haze-free images with a smaller model size and less multiply-accumulate operations. The proposed dehazing method is publicly available at https://github.com/HaoranLiu507/DehazeSNN.

## I. Introduction

Single image dehazing is a vital technique in image processing designed to enhance the visibility and quality of images diminished by atmospheric haze, which can obscure details and distort color accuracy. Traditional dehazing methods extract information from a single hazy image, utilizing models of light propagation and atmospheric scattering to estimate and eliminate haze effects [1]. However, recent advancements in deep learning techniques have demonstrated significantly superior performance, effectively dominating this field. By analyzing contrast and consistency within the hazy image, these algorithms can enhance clear regions while accurately reconstructing obscured details. This process is particularly valuable in applications such as remote sensing [2], photography [3], and computer vision [4], where clarity is crucial for precise analysis and interpretation.

Convolutional Neural Networks (CNNs) [4] and Transformers [5] stand out as the two most prominent

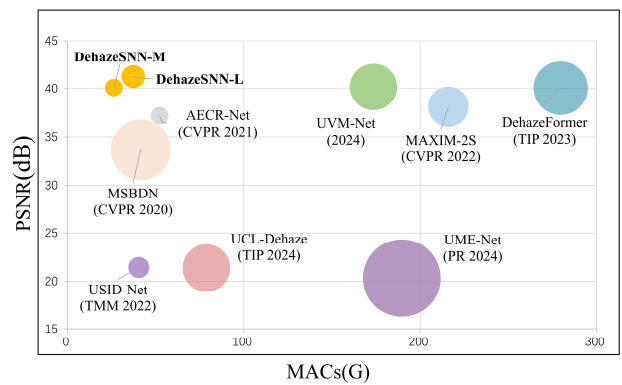

Fig. 1: Comparisons between our DehazeSNN and other state-of-the-art dehazing methods on the RESIDE-ITS set. Circle size is proportional to the number of model parameters.

backbones for image dehazing architectures. CNNs have been widely adopted due to their exceptional ability to capture local features and patterns within images, making them effective for various image processing tasks, including dehazing. They rely on convolutional layers to analyze image segments, providing robust feature extraction. On the other hand, Transformers have gained traction for their capacity to model global dependencies through self-attention mechanisms. This allows them to consider contextual information across the entire image, which is crucial for effectively restoring hazy scenes.

CNN-based models were among the first deep learning methods applied to the field of image dehazing. These models initially built upon traditional algorithms that utilize the atmospheric scattering model, estimating atmospheric light or transmission depth to reconstruct haze-free images through the resolution of physical systems [7]. However, early methods often produced images with high saturation levels and encountered issues such as color distortion and halo artifacts. As research progressed, it became evident that using CNN-based models to directly estimate a haze-free image could yield superior performance [8], resulting in a more streamlined dehazing approach that eliminates the need for complex physical modeling. However, due to the limited size of convolutional kernels, CNN-based models often fall short in capturing long-range dependencies and global image feature extraction. As a result, recent research has increasingly emphasized the use of multi-scale fusion [6, 7], feature pyramids [8, 9], and dilated convolutions [10, 11] to enhance the overall capability of image information processing. These approaches aim to improve the models' ability to gather comprehensive contextual information, leading to better performance in image dehazing tasks.

Transformers-based models utilize attention mechanisms to effectively capture global information within an image. By establishing a direct mapping between hazy and haze-free images, these methods often outperform CNN-based models [15]. However, this remarkable capability for managing long-range dependencies comes at the cost of computationally intensive cross-attention calculations. As a result, Transformers typically require a large number of parameters and an even greater number of Multiply-Accumulate Operations (MACs). To address this computational burden, recent research has focused on optimizing the attention process, leading to the development of models such as DehazeFormer [12] and SwinIR [13], which aim to deliver high performance while reducing computational demands.

To address the limitations of models based on CNNs and Transformers, this study proposes DehazeSNN, which combines a U-Net-like architecture with Spiking Neural Networks (SNNs). The classic U-Net structure is employed to decompose hazy images into multi-scale feature map channels, allowing DehazeSNN to manipulate image features across various scales and effectively capture and restore objects of different sizes. Additionally, we implement an Orthogonal Leaky-Integrate-and-Fire (LIF) module, termed OLIFBlock, to facilitate cross-channel information communication. This design enables DehazeSNN to leverage global image information and long-term dependencies. Consequently, DehazeSNN efficiently processes local features and patterns, akin to CNNs, while managing long-range dependencies similar to Transformers, but with greater efficiency than traditional cross-attention mechanisms.

Moreover, the OLIFBlock inherits significant computational efficiency from SNNs [14], further enhancing the overall performance of DehazeSNN. It achieves state-of-the-art image dehazing results with a considerably smaller model size compared to competing models based on CNNs or Transformers. An intuitive demonstration provided in Fig. 1 shows that both the model size and MACs of DehazeSNN are significantly smaller than those of its competitors. In summary, our contributions are as follows:

- We introduced DehazeSNN, featuring a U-Net-like architecture capable of processing images across different feature channel scales, effectively leveraging rich local features and patterns for image dehazing.
- We developed the OLIFBlock based on Spiking Neural Networks (SNNs), which facilitates access to long-term dependencies while significantly reducing computational burden, marking the first application of SNNs in the field of image dehazing.
- Our DehazeSNN performs favorably against state-of-the-art methods on three benchmark datasets, encompassing both photography and remote sensing dehazing. It achieved high evaluation criteria values while maintaining a very small model size and less MACs.

## II. RELATED WORK

### A. Image Dehazing

Image quality can significantly degrade in hazy weather, adversely affecting digital image processing and interpretation. As a result, many researchers focus on recovering high-quality, clear scenes from hazy images. Prior to the broad adoption of deep learning in computer vision, image dehazing algorithms primarily relied on physical models and prior assumptions, such as the well-known Dark Channel Prior (DCP) [1]. However, these traditional methods struggle to compete with deep learning models due to their reliance on limited prior knowledge of image scenes. Early deep learning approaches sought to estimate the parameters of atmospheric scattering models [7, 8], but more recent studies have shown that end-to-end dehazing methodologies [8], which operate independently of physical models, often achieve superior performance.

The rapid evolution of deep learning architecture has profoundly impacted the mainstream models used in image dehazing. These models have transitioned from Convolutional Neural Networks (CNNs) [15, 16] to Generative Adversarial Networks (GANs) [17], stable diffusion models, and Transformers [18, 19]. In 2020, the introduction of the Vision Transformers (ViT) marked a significant performance boost over previous models, dominating benchmarks across all model sizes [12]. Following this, the Swin Transformers further enhanced image dehazing performance by employing attention mechanisms based on shifted windows [13], enabling it to capture overall image feature dependencies at a considerably lower computational cost.

Meanwhile, CNN-based models have also adapted to address the challenges of managing long-term dependencies. For instance, researchers developed MixDehazeNet [20], which employs multi-scale parallel large convolution kernels to effectively handle uneven hazy distributions. This model outperformed all other state-of-the-art methods in 2023, including those utilizing more advanced architectures like Transformers. However, the reliance on attention mechanisms or large convolution kernels results in larger models with a substantial number of parameters and MACs.

In addition to improving model architecture, many recent studies have focused on enhancing learning methodologies. Orthogonal decoupling contrastive regularization has been introduced to facilitate unpaired image dehazing [21], addressing the challenges of obtaining paired hazy and haze-free images. Similarly, Cong et al. proposed a semi-supervised approach to tackle this issue [22]. Furthermore, the joint learning of depth estimation and dehazing has been implemented [23], enabling a reinforcement learning approach that achieved state-of-the-art performance in 2024. While these advancements have boosted image dehazing capabilities, they have also complicated model architectures and learning processes, resulting in increasingly cumbersome designs.

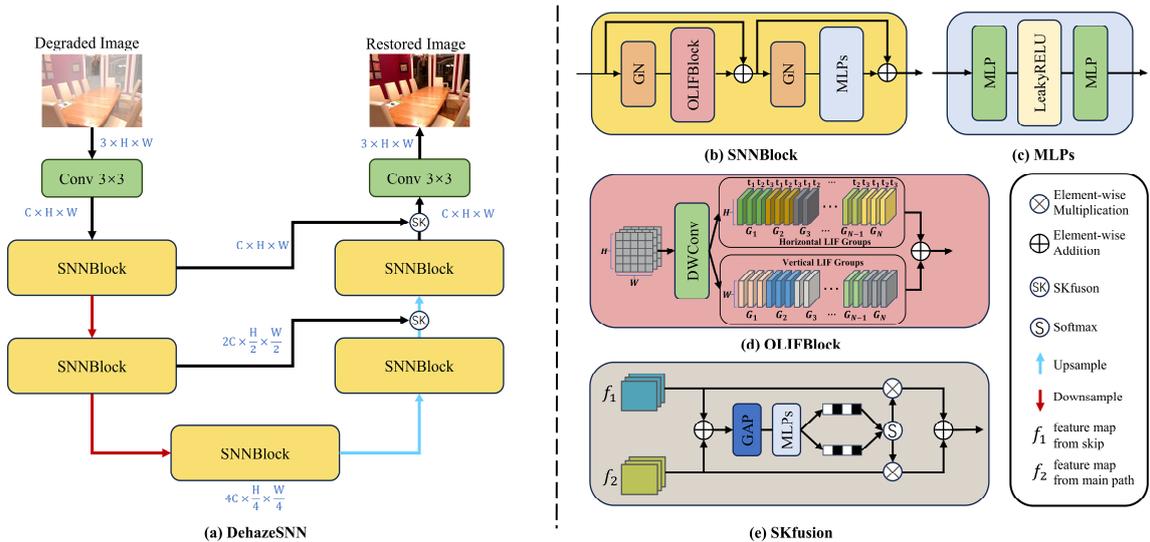

Fig. 2: (a) Overall architecture of the proposed DehazeSNN, which learns image features through a 5-stage U-net. (b) Architecture of Spiking Neural Network Block (SNNBlock), which contains two core parts: (c) MLPs, and (d) Orthogonal Leaky-Integrate-and-Fire Block (OLIFBlock). (e) SKfusion, which serves as a connectivity means to link same-latitude feature maps for information exchange.

## B. Spiking Neural Networks

SNNs are an advanced approach in neuromorphic computing and artificial intelligence, designed to mirror the communication methods of biological neurons through discrete spikes or action potentials [24]. Unlike traditional Artificial Neural Networks (ANNs) that leverage continuous values and activation functions, SNNs encode information in the timing of these spikes. This enables more efficient processing of temporal data and enhances the representation of dynamic environments.

Moreover, SNNs offer substantial reductions in computational costs compared to conventional ANNs, which often depend on extensive MACs [25]. In an SNN, a neuron's internal potential is determined by accumulating the synaptic weights of incoming spikes, thus eliminating the need for multiplication. This approach not only addresses the von Neumann bottleneck but also enables the development of large-scale spiking models on neuromorphic hardware [26].

Due to the challenges posed by the undefined nature of differentiating non-continuous spike generators in SNNs, early models like SpikeProp [27], Tempotron [28], and ReSuMe [29] were difficult to train and had limited applications [29-31]. As a result, recent research has focused on enhancing the learning processes of SNNs. Various solutions have been proposed, including ANN-to-SNN conversion [30], modified backpropagation techniques [32], novel biologically plausible learning rules [33], and continuous spike approximations [34]. Recently, SNNs have found applications in diverse fields such as image classification, segmentation, and robotic control [35]. The architecture and learning mechanisms of SNNs are continually evolving, leading to an expansion of their applicable domains.

## III. METHODOLOGY

In this section, we present DehazeSNN. We begin with an overview of its overall architecture, followed by a detailed explanation of the Spiking Neural Network Block (SNNBlock) and the Orthogonal Leaky-Integrate-and-Fire Block (OLIFBlock). Finally, we describe the SKfusion-based skip connection [12] utilized in DehazeSNN.

### A. DehazeSNN Architecture

DehazeSNN is a 5-stage U-Net [36], as shown in Fig. 2. Our network consists of three parts: shallow feature extraction, deep feature extraction, and image reconstruction. When a hazy image $I \in R^{3 \times H \times W}$ is input into the network, shallow feature extraction is first performed using $3 \times 3$ convolutional layers to generate shallow feature map $F \in R^{C \times H \times W}$. Subsequently, the feature map $F$ is fed into a two-scale encoder-decoder network comprising five SNNblock modules for deep feature extraction, with information transfer between the same scales utilizing SKfusion instead of traditional skip connections. Finally, a $3 \times 3$ convolutional layer is applied to output the reconstructed image $J \in R^{3 \times H \times W}$.

### B. Spiking Neural Network Block

As shown in Fig. 2(b), the SNNBlock consists of two residual sub-blocks: OLIFBlock and Multilayer Perceptron Layers (MLPs). The OLIFBlock adopts an iterative spiking mechanism as a replacement for traditional attention mechanisms, enabling the extraction of local information effectively. It can capture texture features efficiently, facilitating perception of both global and local information. The MLPs, consisting of fully connected layers and activation functions, are responsible for further extracting and mapping the feature information from the OLIFBlock. Additionally, drop path is used to connect the two modules, providing regularization to prevent overfitting.

### C. Orthogonal Leaky-Integrate-and-Fire Module

The Leaky-Integrate-and-Fire (LIF) neuron is a simplified model of neural activity that captures essential features of how real neurons integrate incoming signals and gradually lose potential over time, ultimately firing action potential when a certain threshold is reached. The behavior of a classic LIF neuron can be modeled as follows:

$$\tau \frac{du}{dt} = -u + I, \quad (1)$$

$$o = \begin{cases} 1 & u > V_{th} \\ 0 & u < V_{th} \end{cases}, \quad (2)$$

$$u = \begin{cases} u_{\text{reset}} & u > V_{th} \\ u & u < V_{th} \end{cases}. \quad (3)$$

TABLE I
QUANTITATIVE COMPARISON OF VARIOUS SOTA METHODS ON SYNTHETIC DATASETS

| Methods | | RESIDE-ITS | | RESIDE-OTS | | Overhead | |
|---|---|---|---|---|---|---|---|
| | | PSNR | SSIM | PSNR | SSIM | #Param | MACs |
| DCP [1] | TPAMI'10 | 16.62 | 0.818 | 19.13 | 0.815 | - | - |
| DehazeNet [37] | TIP'16 | 19.82 | 0.821 | 24.75 | 0.927 | 0.009M | 0.581G |
| GridDehazeNet [38] | ICCV'19 | 32.16 | 0.984 | 30.86 | 0.982 | 0.956M | 21.49G |
| MSBDN [39] | CVPR'20 | 33.67 | 0.985 | 33.48 | 0.982 | 31.35M | 41.54G |
| AECR-Net [15] | CVPR'21 | 37.17 | 0.990 | - | - | 2.61M | 52.2G |
| PSD [40] | CVPR'21 | 12.50 | 0.715 | 15.51 | 0.749 | 33.11M | 182.5G |
| MAXIM-2S [41] | CVPR'22 | 38.11 | 0.991 | 34.19 | 0.985 | 14.10M | 108G |
| Dehamer [42] | CVPR'22 | 36.63 | 0.988 | **35.18** | <u>0.986</u> | 132.50M | 24.465G |
| USID-Net [43] | TMM'22 | 21.41 | 0.895 | 23.89 | 0.919 | 3.77M | 40.423G |
| Cycle-SNSPGAN [44] | TITS'22 | 19.13 | 0.852 | 24.28 | 0.925 | 2.358M | 67.145G |
| DehazeFormer [12] | TIP'23 | 40.05 | **0.996** | <u>34.95</u> | 0.984 | 25.44M | 279.7G |
| DDPNet [45] | CVPR'23 | 39.31 | 0.994 | 34.72 | **0.989** | - | - |
| SwinTD-Net [46] | KBS'23 | 36.79 | 0.968 | 32.07 | 0.981 | 22.24M | 1457G |
| UME-Net [47] | PR'24 | 20.3 | 0.704 | 27.83 | 0.953 | 52.229M | 189.62G |
| UVM-Net [48] | arXiv'24 | <u>40.17</u> | **0.996** | 34.92 | 0.984 | 19.25M | 173.55G |
| UCL-Dehaze [49] | TIP'24 | 21.36 | 0.862 | 25.21 | 0.927 | 19.451M | 78.795G |
| DehazeSNN-M | | 40.10 | <u>0.995</u> | - | - | 2.70M | 26.28G |
| DehazeSNN-L | | **41.26** | 0.996 | 33.69 | 0.982 | 4.75M | 37.27G |

TABLE II
QUANTITATIVE COMPARISON ON RESIDE-6K DATASETS.

| Methods | PSNR | SSIM | #Params | MACs |
|---|---|---|---|---|
| DCP [1] | 17.88 | 0.816 | - | - |
| DehazeNet [37] | 21.02 | 0.870 | 0.009M | 0.581G |
| MSBDN [39] | 28.56 | 0.966 | 31.35M | 41.54G |
| FFA-Net [50] | 29.96 | <u>0.973</u> | 4.46M | 287.5G |
| AECR-Net [15] | 28.52 | 0.964 | 2.61M | 52.2G |
| Dehamer [42] | 27.52 | 0.950 | 29.44M | 59.67G |
| SDA-GAN [51] | 18.67 | 0.795 | 19.96M | 72.19G |
| IR-SDE [52] | 28.5 | 0.958 | 5.45M | 41.95G |
| CCA [53] | 29.06 | 0.951 | 4.75M | 306.9G |
| PNE [53] | 29.64 | 0.964 | 4.75M | 306.9G |
| WAE [54] | 25.84 | 0.949 | 12.5M | 0.08G |
| Bi-Dehazing [55] | <u>30.71</u> | **0.975** | 8.054M | - |
| DehazeSNN-M | 30.07 | <u>0.973</u> | 2.70M | 26.28G |
| DehazeSNN-L | **30.77** | **0.975** | 4.75M | 37.27G |

TABLE III
QUANTITATIVE COMPARISON ON RS-HAZE DATASETS.

| Methods | PSNR | SSIM | #Params | MACs |
|---|---|---|---|---|
| DCP [1] | 17.86 | 0.734 | - | - |
| DehazeNet [37] | 23.16 | 0.816 | 0.009M | 0.581G |
| MSBDN [39] | 38.57 | 0.965 | 31.35M | 41.54G |
| AECR-Net [15] | 35.69 | 0.959 | 2.61M | 52.2G |
| Restormer [56] | 32.94 | 0.958 | 26.13M | - |
| M2SCN [57] | 37.75 | 0.950 | - | - |
| Trinity-Net [58] | 32.17 | 0.919 | - | - |
| C2PNET [59] | 34.78 | 0.942 | 7.17M | - |
| DehazeSNN-M | <u>38.95</u> | <u>0.970</u> | 2.70M | 26.28G |
| DehazeSNN-L | **39.14** | **0.971** | 4.75M | 37.27G |

Where $u$ represents the membrane potential of the neuron, $I$ is the input from the upper layers acting as external stimulus, $\tau$ is the time decay coefficient, $o$ is the final output pulse, and $V_{th}$ is the firing threshold for the current neuron.

When $u$ is below the threshold $V_{th}$, the membrane potential decays over time and receives external stimuli, and at this point, the neuron does not emit a pulse signal. When u exceeds the threshold $V_{th}$, the neuron emits a pulse, and $u$ is then reset to the resting membrane potential $u_{\text{reset}}$.

Traditionally, LIF neurons accumulate charge potential over time, which poses challenges for large-scale parallel computation in deep neural networks on GPUs. To effectively integrate LIF neurons into deep neural networks, DehazeSNN accumulates charge potential in the spatial domain by an iterative grouping technique for processing feature maps. It divides the feature map into horizontal and vertical branches, independently integrating information within these branches spatially rather than relying on time-based accumulation (i.e., working in group steps instead of time steps). Additionally, DehazeSNN also replaces LIF binary output pulses with continuous outputs, enabling full-precision outputs that facilitate gradient descent for parameter optimization. The formula is as follows:

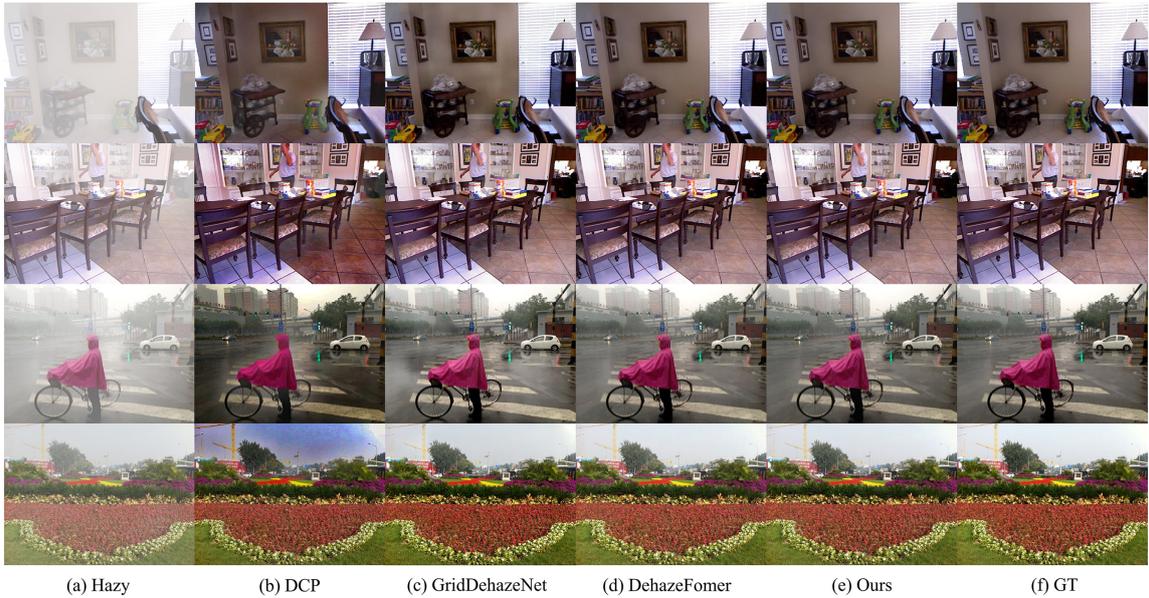

(a) Hazy     (b) DCP     (c) GridDehazeNet     (d) DehazeFomer     (e) Ours     (f) GT

Fig. 3: Qualitative comparison of synthetic hazy images by different methods. The first two rows of images are from the RESIDE- ITS dataset, the last two rows are from RESIDE- OTS dataset.

TABLE IV: MODEL ARCHITECTURE DETAILED.

| Setting | DehazeSNN-M | DehazeSNN-L |
|---|---|---|
| Num. of Blocks | 8, 12, 16, 12, 8 | 8, 16, 32, 16, 8 |
| Embedding Dims | 24, 48, 96, 48, 24 | 24, 48, 96, 48, 24 |
| MLP Ratio | 4, 4, 4, 4, 4 | 4, 4, 4, 4, 4 |
| LIF Init $\tau$ | 0.25 | 0.25 |
| LIF Init $V_{th}$ | 0.25 | 0.25 |

$$y_{g+1} = \sum W^T x, \quad (4)$$

$$u_{g+1} = \tau u_g (1 - o_g) + y_{g+1}, \quad (5)$$

$$o_{g+1} = \begin{cases} 1 & u_{g+1} > V_{th} \\ 0 & u_{g+1} < V_{th} \end{cases}, \quad (6)$$

$$r_{g+1} = \max(u_{t+1}^n, V_{th}). \quad (7)$$

Where $g$ represents the group number. $W$, $x$, and $y$ represent the weight, input, and output, respectively. Different from traditional LIF neurons, OLIFBlock does not directly output variable $o_{g+1}$. It outputs $r_{g+1}$ as the full-precision result at the $g+1$ group step. $o_{t+1}^n$ instead serves as a temporary variable to store the output state at the $g+1$ group step and participates in the next round of iterative calculations.

The detailed processing scheme of the OLIFBlock is illustrated in Fig. 2(d). Initially, the feature map undergoes convolution processing using Depth-Wise Convolution (DWConv). To extract information from various directions and establish long-term dependencies, the feature maps are then processed in parallel through two directional groups: horizontal and vertical branches. In each branch, spatial information is accumulated through LIF calculations of internal neural potential. Finally, the outputs from the two LIF groups are combined, resulting in a feature map that maintains the same size as the original input.

### D. Residual Connection with SKfusion

In the classic U-Net architecture, the fusion of skip branch and main branch is typically achieved through concatenation [36], where the branches are directly joined and subsequently reduced in dimensionality via convolutions. However, this straightforward concatenation often overlooks differences and biases in the feature map branches. To address this issue, we utilize SKfusion for merging information from the two branches. As illustrated in Fig. 2(e), the feature maps from both branches first undergo element-wise addition. The resulting feature map is then processed using Global Average Pooling (GAP) and passed through a MLP module to generate an attention vector for each feature map. These attention vectors act as weights, allowing for a selective focus on relevant information from each branch, all while maintaining low computational cost, making SKfusion a promising alternative to traditional concatenation fusion.

## IV. EXPERIMENTS

We implemented DehazeSNN based on PyTorch [60], and all experiments were trained using only one NVIDIA GeForce RTX 4090 GPU. There are two variants for DehazeSNN, namely DehazeSNN-M and DehazeSNN-L, with specific designs listed in TABLE IV. During the training process, we used the AdamW [61] optimizer, with patch size set to $256 \times 256$, batch sizes of 4 (DehazeSNN-L) and 5 (DehazeSNN-M). The exponential decay rates were set to $\beta_1 = 0.9$ and $\beta_2 = 0.999$. The initial learning rates for the parameters $V_{th}$ and $\tau$ of the OLIFBlock were set to $5 \times 10^{-5}$ and the rest of the network's learning rates were set to $1 \times 10^{-4}$ and gradually decreased to $1 \times 10^{-6}$. We used L1 loss and perceptual loss LPIPS [62] as the combined loss functions.

### A. Datasets and Evaluation Criteria

We evaluated our proposed model on the RESIDE [63] and RS-Haze [12] datasets. The RESIDE dataset is widely utilized in the field of image dehazing as a benchmark ground. It consists of several sub-datasets, each unique in its characteristics and applications. In our experiments, we used three subsets as training data: ITS (13,990 pairs of indoor images), OTS (313,950 pairs of outdoor images), and the mixed dataset RESIDE-6k (6,000 pairs of indoor and outdoor images), and evaluated them on SOTS indoor (500 pairs of images) and SOTS outdoor (500 pairs of images). To evaluate the performance of our method on more diverse hazy conditions, we used the RS-Haze dataset (51,300 pairs of training images and 2,700 pairs of testing images), as the haze in remote sensing images is highly non-uniform. Performance evaluation was done using Peak Signal-to-Noise Ratio (PSNR) and Structural Similarity Index Measure (SSIM). We also used two metrics to quantify computational costs: the number of parameters and MACs.

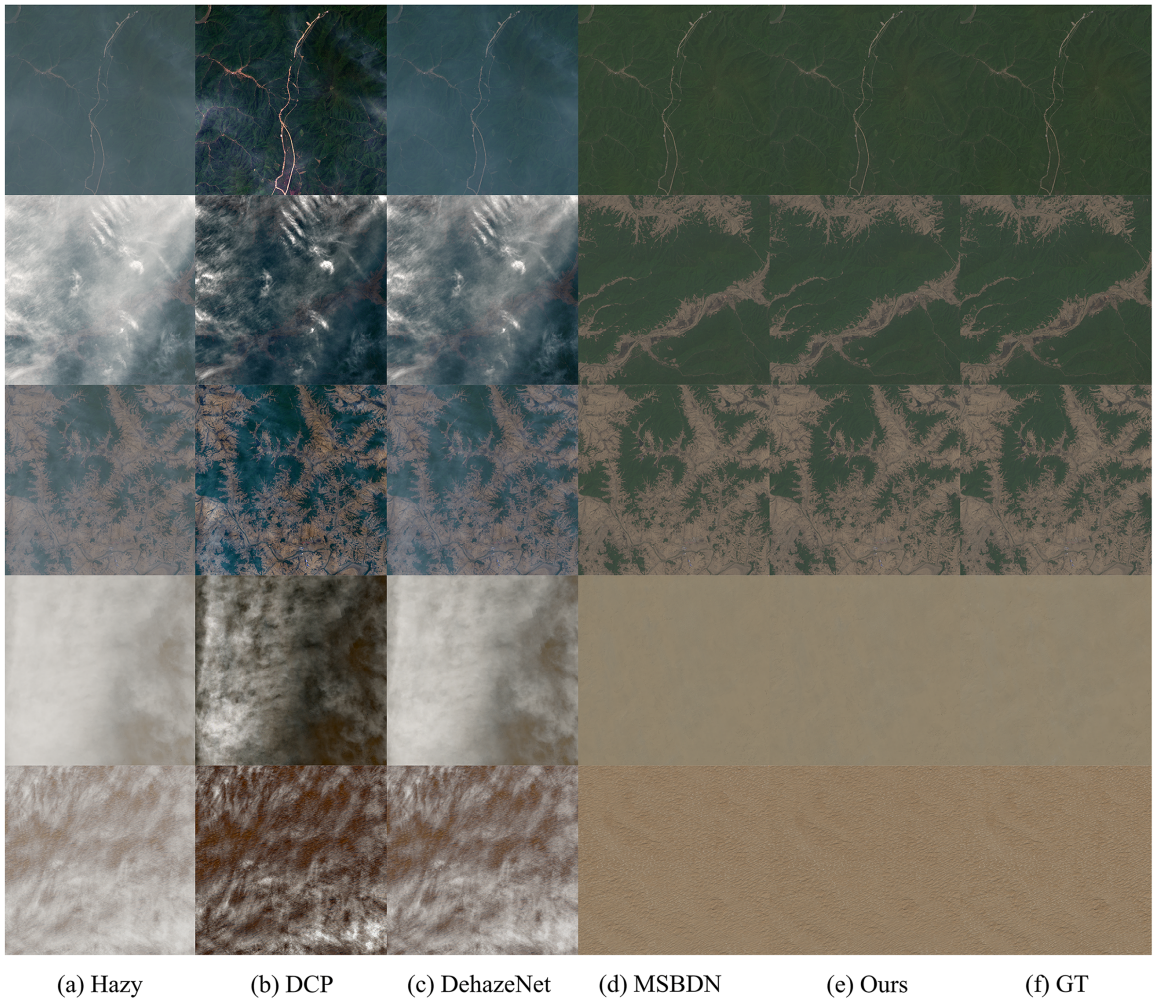

(a) Hazy    (b) DCP    (c) DehazeNet    (d) MSBDN    (e) Ours    (f) GT

Fig. 4: Qualitative comparison of RS hazy images by different methods.

TABLE V:
ABLATION STUDIES ON COMPONENTS OF DEHAZESNN.

| Methods | PSNR | SSIM | #Params | MACs |
|---|---|---|---|---|
| Baseline | 25.69 | 0.938 | 1.77M | 17.52G |
| +SKfusion | 25.70 | 0.939 | 1.76M | 17.38G |
| +OLIFBlock | 29.88 | 0.971 | 2.70M | 26.43 |
| DehazeSNN-M | 30.07 | 0.973 | 2.70M | 26.28G |

### B. Photography Dehazing

We compared DehazeSNN with recent methods in the image dehazing field, and the results are shown in TABLE I and TABLE II. The quantitative results in TABLE I demonstrate that our DehazeSNN-L outperforms all methods on the RESIDE-ITS dataset, not only improving the PSNR value to above 41 but also having significantly lower parameter and computational costs compared to models with PSNR values above 40. This implies that our model is more suitable for tasks focused on lightweight and performance-oriented applications. Additionally, in the RESIDE-OTS dataset, which is characterized by its substantially larger size, a compact model like DehazeSNN struggles to comprehensively capture its features. Consequently, it ranks only among the top performers in the table. Nevertheless, this still highlights the advantage of our model in terms of parameter efficiency and computational performance. Moreover, its results for RESIDE-6K, presented in TABLE II, surpass those of other algorithms, demonstrating that DehazeSNN excels not only on specific features but also exhibits robust feature extraction capabilities across diverse, mixed datasets. Fig. 3 illustrates the visual comparison of DehazeSNN with other dehazing methods. In indoor image dehazing, comparative methods often exhibit artifacts in edge shadow areas, while DehazeSNN does not have these issues and shows better processing results. In outdoor scenes, at the boundary of ground water mist, all comparative methods exhibit varying degrees of residual haze. In comparison, our DehazeSNN generates images that are most faithful to the original, with better recovery of texture details at the boundary.

TABLE VI: ABLATION STUDY ON LOSS FUNCTION

| Loss Function | PSNR | SSIM |
|---|---|---|
| L1 | 28.91 | 0.962 |
| LPIPS | 26.57 | 0.941 |
| 0.5LPIPS+0.5L1 | 30.07 | 0.973 |
| 0.3 LPIPS+0.7L1 | 29.02 | 0.967 |

TABLE VII: ABLATION STUDY ON LIF GROUP NUMBER

| LIF Group number | PSNR | SSIM |
|---|---|---|
| 2 | 29.26 | 0.969 |
| 6 | 28.95 | 0.963 |
| 4 | 30.07 | 0.973 |

*C. Remote Sensing Dehazing*

TABLE III displays the comparison results of our method with others on the RS-Haze dataset. DehazeSNN still achieves good results in handling remote sensing images with non-uniform haze. Our approach, as shown in Fig. 4, produces visual results closer to haze-free real images, demonstrating better performance in color and contrast. Such results are demonstrated objectively in TABLE III, where DehazeSNN significantly outperformed advanced dehazing methods in the field with minimal parameter amount and model size.

*D. Ablation Study*

In this section, we conduct an in-depth analysis based on DehazeSNN-M, elucidating its core components and the impact of related parameter selections. All models in the analysis were trained on the RESIDE-6K dataset and evaluated on the mixed SOTS indoor-outdoor dataset.

**Effectiveness of the components.** We established a U-net-based baseline model, without SKfusion and OLIFBlock. Subsequently, we configured two variants based on the baseline model to verify the effectiveness of these two components of DehazeSNN, as shown in TABLE V. SKfusion does not substantially boost overall performance; however, it offers lower costs than traditional concatenation fusion, positioning it as an effective alternative. The OLIFBlock plays a pivotal role in driving the model's overall performance. Following its integration, performance metrics showed significant improvement, strongly confirming its effectiveness in feature extraction.

**Study of the Loss Function.** DehazeSNN utilizes a weighted combination of L1 Loss and LPIPS Loss as the Loss function, with the formula defined as follows:

$$\mathcal{L}_{total} = \alpha_1 \mathcal{L}_1 + (1-\alpha_1)\mathcal{L}_{LPIPS}. \quad (8)$$

To find the optimal value for the hyperparameter $\alpha_1$, we conducted ablation experiments, with the results shown in TABLE VI, indicating that $\alpha_1 = 0.5$ yields the best performance.

**Research on the LIF group number.** We also explored the optimal hyperparameter $g$, which represents the number of LIF neuron groups. We evaluated the results under various values of $g$, as presented in TABLE VII. The highest accuracy was achieved when $g = 4$. Therefore, our model uniformly adopts $g = 4$ in experiments.

V. CONCLUSION

In this study, we introduce DehazeSNN, a U-Net-like spiking neural network designed for image dehazing. DehazeSNN excels in detailed local feature analysis while effectively managing long-term dependencies and significantly reducing computational overhead. These advantages stem from our innovative OLIFBlock, which employs orthogonal image grouping and the leaky-integrate-and-fire neural activity characteristic to extract and generalize image features across various scales.

Experimental results on both photography and remote sensing dehazing datasets demonstrate that DehazeSNN is highly competitive with state-of-the-art methods, achieving first-tier performance with a minimal model size and a low number of parameters. This indicates the considerable potential of DehazeSNN for application in computationally sensitive dehazing scenarios. Future research will explore the application of DehazeSNN architecture to additional image restoration tasks.

ACKNOWLEDGMENT

This work was supported by the Wenzhou Major Science and Technology Innovation Project (No. ZG2023011) and partially funded by the Natural Science Foundation of Zhejiang Province (No. LZ25F010007).